\documentclass[journal,twoside]{IEEEtran}
\ifCLASSINFOpdf
\else
\fi
\usepackage{verbatim}
\usepackage{amsmath}
\usepackage{amssymb}
\usepackage[slantedGreek]{mathptmx}
\usepackage{threeparttable}
\usepackage{cite}
\usepackage{enumerate}
\usepackage{color}
\usepackage{psfrag}
\usepackage{ifpdf}
\usepackage{multirow}
\usepackage{setspace}
\usepackage{url}
\usepackage{algorithmic}
\usepackage{algorithm}
\usepackage{booktabs}
\usepackage{array}
\usepackage{amsfonts}
\usepackage{graphicx}
\usepackage[numbers,sort&compress]{natbib}
\usepackage[caption=false,font=footnotesize]{subfig}
\usepackage{rotating}
\usepackage{textcomp}
\usepackage{tabularx}
\usepackage{bm}

\newcolumntype{L}[1]{>{\raggedright\arraybackslash}p{#1}}
\newcolumntype{C}[1]{>{\centering\arraybackslash}p{#1}}
\newcolumntype{R}[1]{>{\raggedleft\arraybackslash}p{#1}}

\begin{document}
%
\title{DIMENSION: Dynamic MR Imaging with Both K-space and Spatial Prior Knowledge Obtained via Multi-Supervised Network Training}
%
%
%

\author{Shanshan~Wang,~\IEEEmembership{Member,~IEEE,}
    Ziwen Ke, Huitao Cheng, Sen Jia, Leslie Ying, Hairong Zheng 
    and~Dong Liang,~\IEEEmembership{Senior Member,~IEEE}
\thanks{This work has been submitted to the IEEE for possible publication. Copyright may be transferred without notice, after which this version may no longer be accessible.}
\thanks{This research was partly supported by the National Natural Science Foundation of China (61601450, 61871371, 81830056, 61771463, 61471350), Science and Technology Planning Project of Guangdong Province (2017B020227012), the Basic Research Program of Shenzhen (JCYJ20150831154213680), and Key Laboratory for Magnetic Resonance and Multimodality Imaging
of Guangdong Province (2017YFC0108802). * indicates the corresponding author.}
\thanks{Shanshan~Wang,
	Ziwen Ke contributed equally to this manuscript}
\thanks{S. Wang, Z. Ke, H. Cheng, S. Jia, H. Zheng and D. Liang are with Paul C. Lauterbur Research Center for Biomedical Imaging, Shenzhen Institutes of Advanced Technology, Chinese Academy of Sciences, Shenzhen 518055, China (e-mail: sophiasswang@hotmail.com, ss.wang@siat.ac.cn; dong.liang@siat.ac.cn).}
\thanks{L. Ying is with the Department of Biomedical Engineering and the Department of Electrical Engineering, University at Buffalo, The State University of New York, Buffalo, NY 14260 USA }
}

\maketitle

\begin{abstract}
Dynamic MR image reconstruction from incomplete k-space data has generated great research interest due to its capability in reducing scan time. Nevertheless, the reconstruction problem is still challenging due to its ill-posed nature. Most existing methods either suffer from long iterative reconstruction time or explore limited prior knowledge. This paper proposes a dynamic MR imaging method with both k-space and spatial prior knowledge integrated via multi-supervised network training, dubbed as DIMENSION. Specifically, the DIMENSION architecture consists of a frequential prior network for updating the k-space with its network prediction and a spatial prior network for capturing image structures and details. Furthermore, a multi-supervised network training technique is developed to constrain the frequency domain information and reconstruction results at different levels. The comparisons with classical k-t FOCUSS, k-t SLR, L+S and the state-of-the-art CNN-based method on in vivo datasets show our method can achieve improved reconstruction results in shorter time.
\end{abstract}


%
\IEEEpeerreviewmaketitle

\section{Introduction}
%
%
%
%
\IEEEPARstart{D}{ynamic} MR imaging is a non-invasive imaging technique which could provide both spatial and temporal information for the underlying anatomy. Nevertheless, both physiological and hardware constraints have made it suffer from slow imaging speed or long imaging time, which may lead to patients’ discomfort or sometimes cause severe motion artifacts. Therefore, it is of great necessity to accelerate MR imaging.

To accelerate dynamic MR scan, there have been three main directions of efforts, namely in developing physics based fast imaging sequences \cite{kaiser1989mr}, hardware based parallel imaging techniques \cite{sodickson1997simultaneous} and signal processing based MR image reconstruction methods from incomplete k-space data. Our specific focus here is the undersampled MR image reconstruction, which requires prior information to solve the aliasing artifacts caused by the violation of the Nyquist sampling theorem. Specifically, the reconstruction task is normally formulated as solving an optimization problem with two terms i.e. data fidelity and prior regularization. Popular prior information includes sparsity, which prompts image to be sparsely represented in a certain transform domain while being reconstructed from incoherently undersampled k-space data. These techniques are well-known as compressed sensing MRI (CS-MRI) \cite{donoho2006compressed, lustig2007sparse}. 
For example, k-t FOCUSS \cite{jung2007improved} takes advantage of the sparsity of x-f support to reconstruct x-f images from the undersampled k-t space. It encompasses the celebrated k-t BLAST and k-t SENSE \cite{tsao2003k} as special cases. And k-t ISD \cite{liang2012k} incorporates additional information on the support of the dynamic image in x-f space based on the theory of CS with partially known support. DLTG \cite{caballero2014dictionary} can learn redundancy in the data via an auxiliary constraint on temporal gradients (TG) sparsity. Wang et al \cite{wang2014compressed} employs a patch-based 3-D spatiotemporal dictionary for sparse representations of dynamic image sequence. Besides, low-rank is also a prior regularization. It can use low-rank and incoherence conditions to complete missing or corrupted entries of a matrix. A typical example on low-rank is L+S \cite{otazo2015low}, where the nuclear norm is used to enforce low rank in L and the $l_1$ norm is used to enforce sparsity in S. And k-t SLR \cite{lingala2011accelerated} exploits the correlations in the dynamic imaging dataset by modeling the data to have a compact representation in the Karhunen Louve transform (KLT) domain. These methods have made great progresses in dynamic imaging and achieved improved results. Nevertheless, these methods only draw prior knowledge from limited samples. Furthermore, the reconstruction is iterative and sometimes time-consuming. 

On the other hand, deep learning has shown great potential in accelerating MR imaging. There have been quite a few newly proposed methods, which can be roughly categorized into two types, model-based unrolling methods \cite{sun2016deep,hammernik2018learning, knoll2018assessment} and end-to-end learning methods \cite{wang2016accelerating, kwon2017parallel, han2018deep, zhu2018image, eo2018kiki,  sun2018compressed, quan2018compressed, schlemper2018deep, qin2018convolutional, wang2018image, aggarwal2018modl}. The model based unrolling methods are to formulate the iterative procedure of traditional optimization algorithms to network learning. They adaptively learn all the parameters of regularization terms and transforms in the model by network training. For example, in VN-Net \cite{hammernik2018learning}, generalized compressed sensing reconstruction formulated as a variational model is embedded in an unrolled gradient descent scheme. ADMM-Net \cite{sun2016deep} is defined over a data flow graph, which is derived from the iterative procedures in Alternating Direction Method of Multipliers (ADMM) algorithm for optimizing a CS-based MRI model. The other type utilizes the big data information to learn a network that map between the undersampled and fully sampled data pairs. Wang et al. \cite{wang2016accelerating} train a deep convolutional neural
network (CNN) to learn the mapping relationship between undersampled brain MR images and fully sampled brain MR images. AUTOMAP \cite{zhu2018image} learns a mapping between the sensor and the image domain from an appropriate training data. Despite all the successes, there are only two works that specifically apply to dynamic MR imaging \cite{schlemper2018deep, qin2018convolutional}. Both of these two works use a cascade of neural networks to learn the mapping between undersampled and fully sampled cardiac MR images, where a deep cascaded of convolutional neural network (DC-CNN) is designed in \cite{schlemper2018deep} and a convolutional recurrent neural network (CRNN) is proposed in \cite{qin2018convolutional}. Both works make great contributions to dynamic MR imaging. Nevertheless, the reconstruction results can still be improved based on two observations. Firstly, they have adopted a single supervised loss function which only considers the fidelity between the final output and the ground truth. The intermediate results have not been utilized for the supervision. Furthermore, their entire networks are built in the spatial domain where only data-consistency layers are considered for the direct k-space correction and no other k-space information is utilized. However, previous research has shown that the combination of frequential domain networks and spatial domain networks is superior to single-domain CNN \cite{eo2018kiki}. There are still more valuable prior knowledge regarding k-space and different levels of reconstruction to be utilized for accurate MR image reconstruction. 

In this work, we propose a \textbf{D}ynam\textbf{I}c \textbf{M}R imaging method with both k-spac\textbf{E} a\textbf{N}d \textbf{S}patial prior knowledge integrated via mult\textbf{I}-supervised netw\textbf{O}rk trai\textbf{N}ing, dubbed as DIMENSION. The improvements are mainly reflected in the cross-domain network structures and the multi-supervised loss function strategy. Our contributions could be summarized as follows:
\begin{enumerate}
	\item In the present study, a dynamic MR imaging method with both k-space and spatial prior knowledge integrated via multi-supervised network training is proposed, which can combine frequential domain and spatial domain information sufficiently. 
	\item We propose a multi-supervised loss function strategy, which can simultaneously constrain the frequency domain information and the spatial domain information. Such loss function strategy is designed to get better network predicted k-space with the frequency domain learning, and can also prompt the reconstruction results at different levels in the spatial domain learning to be closer to the fully sampled MR images. 
	\item Experimental results show that the proposed method is superior to conventional CS-based methods such as k-t FOCUSS, k-t SLR and L+S, as well as the state-of-the-art CNN-based method, DC-CNN. These demonstrate the effectiveness of the cross-domain learning and the multi-supervised loss function strategy in cardiac MR imaging.	
\end{enumerate}

\section{Methodology}

\subsection{CS-MRI and CNN-MRI}
According to compressed sensing (CS) \cite{donoho2006compressed, lustig2007sparse}, MR images with a sparse representation in some transform domain can be reconstructed from randomly undersampled k-space data. Let $\textbf{S}\in \mathbb{C}^{N_xN_yN_t}$ represent a complex-valued dynamic MR image. The problem can be described by the following formula:
\begin{equation}
\label{eq_1}
\textbf{K}_u = \textbf{F}_u\textbf{S}+\textbf{e}
\end{equation}
where $\textbf{K}_u\in \mathbb{C}^{N_xN_yN_t}$ is the undersampled measurements in k-space and the unsampled points are filled with zeros. $\textbf{F}_u$ is an undersampled Fourier encoding matrix, and $\textbf{e}\in\mathbb{C}^{N_xN_yN_t}$ is the acquisition noise. We want to reconstruct $\textbf{S}$ by solving the inverse problem of Eq. \ref{eq_1}. However, the inverse problem is ill-posed, resulting in that the reconstruction is not unique. In order to reconstruct $\textbf{S}$, we constrain this inverse problem by adding some prior knowledge and solve the following optimization problem:
\begin{equation}
\label{eq_2}
\min_\textbf{S} \frac{1}{2} ||\textbf{F}_u\textbf{S}-\textbf{K}_u||_2^2+\lambda\mathcal{R}(\textbf{S})
\end{equation}
The first term is the data fidelity, which ensures that the k-space of reconstruction is consistent with the actual measurements in k-space. The second term is often referred to as the prior regularization. In the methods of CS, $\mathcal{R}(\textbf{S})$ is usually a sparse prior of $\textbf{S}$ in some transform domains, e.g. finite difference, wavelet transform and discrete cosine transformation.

In CNN-based methods, $\mathcal{R}(\textbf{S})$ is a CNN prior of $\textbf{S}$ , which force $\textbf{S}$ to match the output of the networks:
\begin{equation}
\label{eq_3}
\min_\textbf{S} \frac{1}{2} ||\textbf{F}_u\textbf{S}-\textbf{K}_u||_2^2+\lambda||\textbf{S}-f_{CNN}(\textbf{S}_u|\bm{\theta})||_2^2
\end{equation}
where $\textbf{S}_u$ is the undersampled image and $f_{CNN}(\textbf{S}_u|\bm{\theta})$ is the output of the  networks under the parameters $\bm{\theta}$. The training process of the networks is to find the optimal parameters $\bm{\theta}^*$. Once the network are trained, the networks' output $f_{CNN}(\textbf{S}_u|\bm{\theta}^*)$ is the reconstruction we want.

\subsection{The Proposed Method}

\subsubsection{The Proposed DIMENSION Network}
\begin{figure*}[!t]
	\centering
	\subfloat{\includegraphics[width=0.9\linewidth]{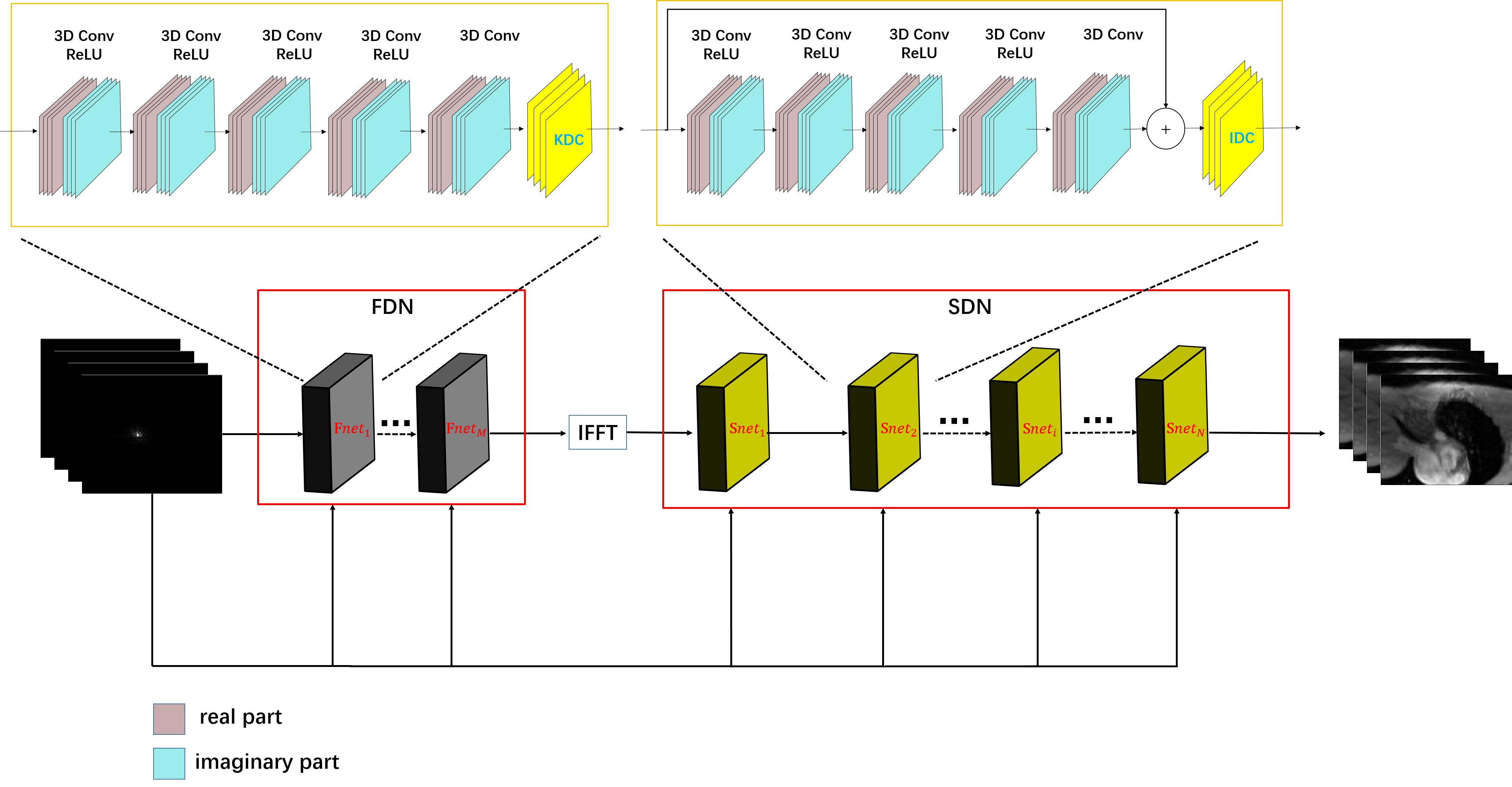}
		\label{Subnetworks}}
	
	\caption{The proposed DIMENSION network architecture for cardiac MR reconstruction.}
	\label{networks}
\end{figure*}

In this work, we propose a convolutional neural network termed as DIMENSION for cardiac MR images reconstruction shown in Fig.\ref{networks}. The DIMENSION network consists of two main parts: a frequency domain network for updating the k-space with its network prediction termed as FDN and a spatial domain network term as SDN, which is used to extract high-level features of images. The FDN and the SDN are connected by a Fourier inversion (see Inverse Fast Fourier Transform (IFFT) in Fig. \ref{networks}).

Specifically, the FDN consists of $M$ frequency domain blocks ${\rm \textbf{F}}_{net_m},\ m=1,\dots, M$. Each block contains $L$ 3D convolutional layers and a k-space domain data consistency (KDC) layer. The forward-pass starts from the undersampled k-space $\textbf{K}_u$. The forward-pass equations of the first block $(m=1)$ in the FDN could be described as:
\begin{equation}
\begin{cases}
\textbf{K}_{\textbf{F}_{net_1}}^1=\sigma(\textbf{W}_{\textbf{F}_{net_1}}^1*\textbf{K}_u+\textbf{b}_{\textbf{F}_{net_1}}^1) \\
\ \ \ \ \ \ \ \dots \\
\textbf{K}_{\textbf{F}_{net_1}}^{l+1}=\sigma(\textbf{W}_{\textbf{F}_{net_1}}^{l+1}*\textbf{K}_{\textbf{F}_{net_1}}^l+\textbf{b}_{\textbf{F}_{net_1}}^{l+1}) \\
\ \ \ \ \ \ \ \dots \\
\textbf{K}_{\textbf{F}_{net_1}}^L=\textbf{W}_{\textbf{F}_{net_1}}^L*\textbf{K}_{\textbf{F}_{net_1}}^{L-1}+\textbf{b}_{\textbf{F}_{net_1}}^L \\
\textbf{K}_{\textbf{F}_{net_1}}^{DC}={\rm KDC}(\textbf{K}_{\textbf{F}_{net_1}}^L) 
\end{cases}
\end{equation}
The forward-pass equations of the later blocks $(m=2, \dots, M)$ in the FDN could be described as:
\begin{equation}
\begin{cases}
\textbf{K}_{\textbf{F}_{net_m}}^1=\sigma(\textbf{W}_{\textbf{F}_{net_m}}^1*\textbf{K}_{\textbf{F}_{net_1}}^{DC}+\textbf{b}_{\textbf{F}_{net_m}}^1) \\
\ \ \ \ \ \ \ \dots \\
\textbf{K}_{\textbf{F}_{net_m}}^{l+1}=\sigma(\textbf{W}_{\textbf{F}_{net_m}}^{l+1}*\textbf{K}_{\textbf{F}_{net_m}}^l+\textbf{b}_{\textbf{F}_{net_m}}^{l+1}) \\
\ \ \ \ \ \ \ \dots \\
\textbf{K}_{\textbf{F}_{net_m}}^L=\textbf{W}_{\textbf{F}_{net_m}}^L*\textbf{K}_{\textbf{F}_{net_m}}^{L-1}+\textbf{b}_{\textbf{F}_{net_m}}^L \\
\textbf{K}_{\textbf{F}_{net_m}}^{DC}={\rm KDC}(\textbf{K}_{\textbf{F}_{net_m}}^L) 
\end{cases}
\end{equation}
where the KDC is defined as the following equation:
\begin{equation}
\label{kc_dc}
\textbf{K}_{\textbf{F}_{net_m}}^{DC}=
\begin{cases} 
\frac{\textbf{K}_{\textbf{F}_{net_m}}^L+\lambda\textbf{K}_u(k_x,k_y)}{1+\lambda},  & \mbox{if }(k_x,k_y)\in \Omega \\
\textbf{K}_{\textbf{F}_{net_m}}^L, & \mbox{if } (k_x,k_y)\notin \Omega
\end{cases}
\end{equation}
$\textbf{W}_{\textbf{F}_{net_m}}^l$, $\textbf{b}_{\textbf{F}_{net_m}}^l$ are the $l$-th convolution filters, biases in the $m$-th block of the FDN respectively with $l=1, \dots, L,\ m=1, \dots, M$. $\textbf{K}_{\textbf{F}_{net_m}}^l$ denotes the output of the $l$-th convolutional layer in the $m$-th block in the FDN. Each convolutional layer is followed by an activation function $\sigma$ for nonlinearity except for the last layer, which projects the extracted features to the k-space domain. After the convolution operation, the k-space domain data consistency is utilized to correct network predicted k-space with the actual sampled k-space $\textbf{k}_u$ as shown in Eq. \ref{kc_dc}, where $\textbf{K}_{\textbf{F}_{net_m}}^{DC}$ denotes the correction of $\textbf{K}_{\textbf{F}_{net_m}}^L$. The set consisting of indices of sampled in k-space is defined as $\Omega$. If k-space indices $(k_x,k_y)$ is in set $\Omega$, the $\textbf{K}_{\textbf{F}_{net_m}}^L$ would be corrected with the actual sampled k-space $\textbf{k}_u$. $\lambda$ is used to control the degree of the data consistency. If $\lambda \to \infty$, we replace the unsampled points directly with the actual sampled points.

The final output of the FDN is $\textbf{K}_{\textbf{F}_{net_M}}^{DC}$. Then the inverse Fourier transform of the $\textbf{K}_{\textbf{F}_{net_M}}^{DC}$ is performed to obtain the MR image, which is also the input of the SDN:
\begin{equation}
\label{Ic}
\textbf{S}_0={\rm IFFT}(\textbf{K}_{\textbf{F}_{net_M}}^{DC})
\end{equation}

The SDN consists of $N$ image-domain blocks ${\rm \textbf{S}}_{net_n},\ n=1, \dots, N$, each of which contains $L$ 3D convolutional layers, a residual connection \cite{he2016deep} and an image-domain data consistency layer (IDC). The forward-pass equations of the first block $(n=1)$ in the SDN are: 
\begin{equation}
\begin{cases}
\textbf{S}_{\textbf{S}_{net_1}}^1=\sigma(\textbf{W}_{\textbf{S}_{net_1}}^1*\textbf{S}_0+\textbf{b}_{\textbf{S}_{net_1}}^1) \\
\ \ \ \ \ \ \ \dots \\
\textbf{S}_{\textbf{S}_{net_1}}^{l+1}=\sigma(\textbf{W}_{\textbf{S}_{net_1}}^{l+1}*\textbf{S}_{\textbf{S}_{net_1}}^l+\textbf{b}_{\textbf{S}_{net_1}}^{l+1}) \\
\ \ \ \ \ \ \ \dots \\
\textbf{S}_{\textbf{S}_{net_1}}^L=\textbf{W}_{\textbf{S}_{net_1}}^L*\textbf{S}_{\textbf{S}_{net_1}}^{L-1}+\textbf{b}_{\textbf{S}_{net_1}}^L \\
\textbf{S}_1=\textbf{S}_0+\textbf{S}_{\textbf{S}_{net_1}}^L \\
\hat{\textbf{S}}_{1}={\rm IDC}(\textbf{S}_1) 
\end{cases}
\end{equation}
The forward-pass equations of the later blocks $(n=2, \dots, N)$ in the SDN could be described as:
\begin{equation}
\begin{cases}
\textbf{S}_{\textbf{S}_{net_n}}^1=\sigma(\textbf{W}_{\textbf{S}_{net_n}}^1*\hat{\textbf{S}}_{n-1}+\textbf{b}_{\textbf{S}_{net_n}}^1) \\
\ \ \ \ \ \ \ \dots \\
\textbf{S}_{\textbf{S}_{net_n}}^{l+1}=\sigma(\textbf{W}_{\textbf{S}_{net_n}}^{l+1}*\textbf{S}_{\textbf{S}_{net_n}}^l+\textbf{b}_{\textbf{S}_{net_n}}^{l+1}) \\
\ \ \ \ \ \ \ \dots \\
\textbf{S}_{\textbf{S}_{net_n}}^L=\textbf{W}_{\textbf{S}_{net_n}}^L*\textbf{S}_{\textbf{S}_{net_n}}^{L-1}+\textbf{b}_{\textbf{S}_{net_n}}^L \\
\textbf{S}_n=\hat{\textbf{S}}_{n-1}+\textbf{S}_{\textbf{S}_{net_n}}^L \\
\hat{\textbf{S}}_{n}={\rm IDC}(\textbf{S}_n) 
\end{cases}
\end{equation}
where the IDC is defined as the following equations:
\begin{equation}
\label{S_dc}
\begin{cases}
\textbf{k}_{\textbf{S}_n}={\rm FFT}(\textbf{S}_n) \\
\hat{\textbf{k}}_{\textbf{S}_n}=
\begin{cases} 
\frac{\textbf{k}_{\textbf{S}_n}(k_x,k_y)+\lambda\textbf{k}_u}{1+\lambda},  & \mbox{if }(k_x,k_y)\in \Omega \\
\textbf{k}_{\textbf{S}_n}(k_x,k_y), & \mbox{if } (k_x,k_y)\notin \Omega
\end{cases} \\
\hat{\textbf{S}}_n={\rm IFFT}(\hat{\textbf{k}}_{\textbf{S}_n})
\end{cases}
\end{equation}
$\textbf{W}_{\textbf{S}_{net_n}}^l$, $\textbf{b}_{\textbf{S}_{net_n}}^l$ are the $l$-th convolution filters, biases in the $n$-th block of the SDN respectively with $l=1, \dots,L\ ,n=1,\dots,N$. $\textbf{S}_{\textbf{S}_{net_n}}^l$ denotes the output of the $l$-th convolutional layer in the $n$-th block. Like FDN, each convolutional layer is nonlinear with the activation function $\sigma$ except for the last convolutional layer, which projects the extracted spatial domain features to the image domain. After convolution operation, a residual connection is followed, which sums the output of each block with its input. $\textbf{S}_n$ is the result of the residual learning. Then the image-domain data consistency (IDC) is performed on $\textbf{S}_n$ to obtain the corrected $\hat{\textbf{S}}_{n}$ as shown in Eq. \ref{S_dc}. There are three steps, which respectively represents the fast Fourier transform (FFT), k-space correction, and inverse fast Fourier transform (IFFT). It’s noticeable that for the KDC layer, no frequency domain and spatial domain transformations are necessary, because the output is already a prediction of k-space. Through the above processes, we can get the reconstruction results $\hat{\textbf{S}}_1, \dots, \hat{\textbf{S}}_N$ of the different blocks. $\hat{\textbf{S}}_1, \dots, \hat{\textbf{S}}_N$ come from different network depths. So the reconstruction results are at different levels. For example, $\hat{\textbf{S}}_1$ comes from the shallowest block, so the reconstruction level is the lowest. $\hat{\textbf{S}}_N$ is the final output of the SDN, which is at the highest reconstruction level. 

In \cite{schlemper2018deep}, a DC-CNN model has been proposed to reconstruct undersampled cardiac MR images. The DC-CNN model consists of $N$ image-domain blocks, each of which contains $N$ convolutional layers, a residual connection and an image-domain data consistency layer (IDC). Despite favorable reconstruction quality has been achieved, there are still more valuable prior knowledge regarding k-space can be learned and utilized for improving MR image reconstruction. The DIMENSION model introduces k-space learning into the networks, so that the networks can not only extract features in the spatial domain, but also make better use of k-space prior knowledge.
 
\subsubsection{The Proposed Multi-Supervised Loss}
\begin{figure*}[!t]
	\centering
	\subfloat{\includegraphics[width=0.9\linewidth]{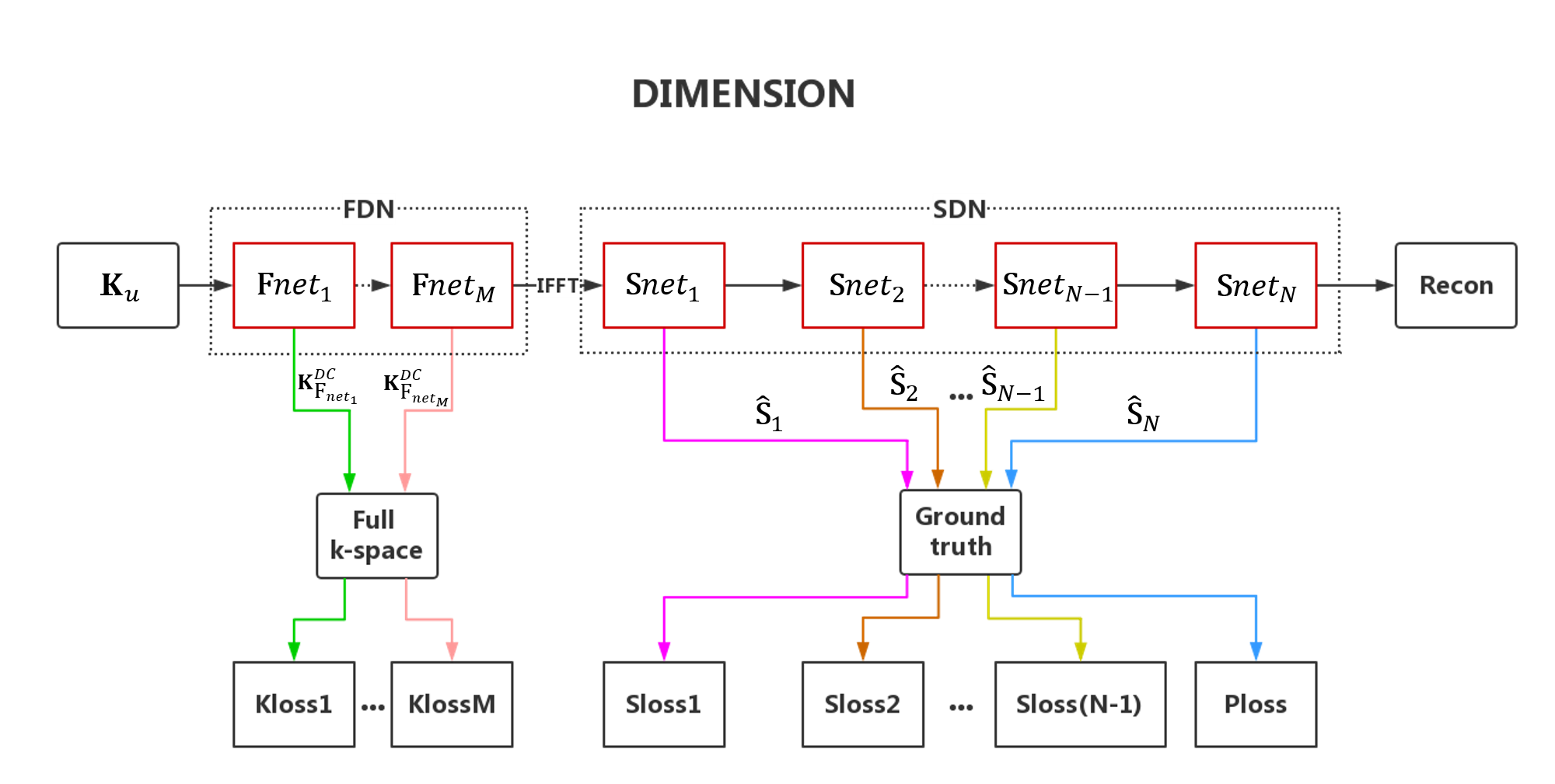}
		\label{Subloss}}
	
	\caption{The multi-supervised loss in DIMENSION networks.}
	\label{loss}
\end{figure*}

In this work, we introduce a multi-supervised loss function strategy. The multi-supervised loss functions are composed of primary loss, k-space loss and spatial loss shown in Fig. \ref{loss}. In the usual CNN-MRI models, only the primary loss exists, which measures the distance between the reconstruction and the ground truth. In this work, the k-space loss and the spatial loss are proposed to constrain the frequency domain information and reconstruction results at different levels. Let $\textbf{K}_f$ be the fully sampled k-space. The k-space loss can be expressed as the formula:
\begin{equation}
\label{eq_6}
\mathrm{Kloss} = \sum_{m=1}^{M}\alpha_m||\textbf{K}_f-\textbf{K}_{\textbf{F}_{net_m}}^{DC}||_2^2
\end{equation}
And the spatial loss can be expressed as the formula:
\begin{equation}
\label{eq_7}
\mathrm{Sloss}
= \sum_{n=1}^{N-1}\beta_n||\textbf{S}-\hat{\textbf{S}}_n||_2^2
\end{equation}
where $\alpha_m$ and $\beta_n$ are the respective nonnegative weights of each loss. The primary loss is the mean square error (MSE) between the final reconstruction ($\hat{\textbf{S}}_N$) and corresponding fully sampled image $\textbf{S}$:
\begin{equation}
\label{eq_8}
\mathrm{Ploss}
= ||\textbf{S}-\hat{\textbf{S}}_N||_2^2
\end{equation}
Finally, the total training loss functions consist of these three terms:
\begin{equation}
\label{eq_9}
\mathrm{Tloss}
= ||\textbf{S}-\hat{\textbf{S}}_N||_2^2+\sum_{m=1}^{M}\alpha_m||\textbf{K}_f-\textbf{K}_{\textbf{F}_{net_m}}^{DC}||_2^2+\sum_{n=1}^{N-1}\beta_n||\textbf{S}-\hat{\textbf{S}}_n||_2^2
\end{equation}
The weights in Eq. \ref{eq_9} will be discussed in the later section. To facilitate the better understanding of this model, we provide an intuitive explanation of the k-space loss and the spatial loss.

The k-space loss: The k-space points can be divided into two subsets: if a point $(k_x,k_y)$ has already been sampled, $(k_x,k_y)$ is a member of set $\Omega$, otherwise $(k_x,k_y)\notin \Omega$. The purpose of the FDN is to update the k-space with its network prediction. The quality of prediction directly affects the image domain feature extraction in the SDN and the final reconstruction. Therefore, it is necessary to enhance the fidelity between the network predicted k-space and the fully sampled k-space. The combination of the KDC and the k-space loss can improve the quality of the predicted k-space. If $(k_x,k_y)\in \Omega$, the KDC layer is good at ensuring that the predicted k-space is consistent with the actual sampled k-space. If $(k_x,k_y)\notin \Omega$, the k-space loss can make the unsampled k-space as close as possible to the fully sampled k-space. 

The spatial loss: the outputs of the spatial blocks from different network depths can be regarded as the reconstruction at different levels. Then the final output of the entire network can be seen as the final level reconstruction. However, previous studies only make constraints on the final level reconstruction and use them as the supervision of the entire networks. Those approaches do not take full advantage of reconstruction information at other levels. Here, we put forward a spatial loss, which can constrain the reconstruction at different levels to be closer to the ground truths. As can be seen in Eq. \ref{eq_9}, reconstruction at each level contributes to the final result. Furthermore, the spatial loss is excepted to alleviate the gradient diffusion problem. Because the new supervisions introduce new gradients, which could avoid the situation that the gradient back-propagates too far and diffuse. 

\section{\textcolor{black}{EXPERIMENTAL RESULTS}}
\subsection{Setup}
\subsubsection{Data acquisition}
We collected 101 fully sampled cardiac MR data using 3T scanner (SIMENS MAGNETOM Trio) with T1-weighted FLASH sequence. Written informed consent was obtained from all human subjects. Each scan contains a single slice FLASH acquisition with 25 temporal frames. The following parameters were used for FLASH scans: FOV $330 \times 330$ mm, acquisition matrix  $192 \times 192$, slice thickness = 6 mm, TR = 3 ms, TE = 50 ms and 24 receiving coils. The raw multi-coil data of each frame was combined by adaptive coil combine method \cite{walsh2000adaptive} to produce a single-channel complex-valued image. We randomly selected 90\% of the entire dataset for training and 10\% for testing. Deep learning has a high demand for data volume \cite{lecun2015deep}. Therefore, some data augmentation strategies have been applied. We shear the original images along the $x, y$ and $t$ direction. The sheared size is $117 \times 120 \times 6 \ (x \times y \times t)$, and the stride along the three directions is 7, 7 and 5 respectively. Finally, we obtained 17500 3D complex-valued cardiac MR data with the size of $117 \times 120 \times 6$.

For each frame, the original k-space was retrospectively undersampled with 6 ACS lines. Specifically, we fully samples frequency-encodes (along $k_x$) and randomly undersamples the phase encodes (along $k_y$) according to a zero-mean Gaussian variable density function \cite{jung2007improved}. 

\subsubsection{Network training}
For network training, we divide each data into two channels, where the channels store real and imaginary parts of the data. So the inputs of the network are undersampled k-spaces $\mathbb{R}^{2N_xN_yN_t}$ and the outputs are reconstruction images $\mathbb{R}^{2N_xN_yN_t}$. In this work, we focus on a D5C5 model, which works pretty well for the DC-CNN model. The D5C5 model consists of five blocks (C5) and each block has five convolutional layers (D5). In order to simplify the parameters $\bm{\theta}$ and make a fair comparison with the D5C5 model, the FDN contains one frequency domain block ($M=1$) and the SDN consists of four image-domain blocks ($N=4$). Every block contains five convolutional layers ($L=5$). Therefore, both the proposed model and the D5C5 model have 25 convolutional layers in total. The details of the FDN and the SDN are shown in Table. \ref{K5} and \ref{D5C4} respectively. He initialization \cite{he2015delving} was used to initialize the network weights. Rectifier Linear Units (ReLU) \cite{glorot2011deep} were selected as the nonlinear activation functions. The mini-batch size was 20. The exponential decay learning rate \cite{zeiler2012adadelta} was used in all CNN-based experiments and the initial learning rate was set to 0.0001 with a decay of 0.95. All the models were trained by the Adam optimizer \cite{kingma2014adam} with parameters $\beta_1=0.9, \beta_2=0.999$ and $\epsilon=10^{-8}$.
\begin{table*}[!t]
	\renewcommand{\arraystretch}{1.1}
	\renewcommand\tabcolsep{3pt}
	\caption{\textcolor{black}{The parameters setting of each block in the FDN.}}
	\label{K5}
	\centering
	\textcolor{black}{\begin{tabular}{|c|c|c|c|c|c|c|} \hline
			Layer of K5 & Input size & Number of filter & Filter size & Stride & Activation  & Output \\ \hline
			Complex conv1 & 117*120*6*2 & 64 & 3*3*3 & [1, 1, 1] & ReLU & 117*120*6*64\\
			\hline
			Complex conv2 & 117*120*6*64 & 64 & 3*3*3 & [1, 1, 1] & ReLU & 117*120*6*64\\
			\hline
			Complex conv3 & 117*120*6*64 & 64 & 3*3*3 & [1, 1, 1] & ReLU & 117*120*6*64\\
			\hline
			Complex conv4 & 117*120*6*64 & 64 & 3*3*3 & [1, 1, 1] & ReLU & 117*120*6*64\\
			\hline
			Complex conv5 & 117*120*6*64 & 2 & 3*3*3 & [1, 1, 1] & None & 117*120*6*2\\
			\hline
			Kspace data consistency & 117*120*6*2 & / & / & / & / & 117*120*6*2\\
			\hline
	\end{tabular}}
\end{table*}
\begin{table*}[!t]
	\renewcommand{\arraystretch}{1.1}
	\renewcommand\tabcolsep{3pt}
	\caption{\textcolor{black}{The parameters setting of each block in the SDN.}}
	\label{D5C4}
	\centering
	\textcolor{black}{\begin{tabular}{|c|c|c|c|c|c|c|} \hline
			Layer of D5C4 & Input size & Number of filter & Filter size & Stride & Activation  & Output \\ \hline
			Complex conv1 & 117*120*6*2 & 64 & 3*3*3 & [1, 1, 1] & ReLU & 117*120*6*64\\
			\hline
			Complex conv2 & 117*120*6*64 & 64 & 3*3*3 & [1, 1, 1] & ReLU & 117*120*6*64\\
			\hline
			Complex conv3 & 117*120*6*64 & 64 & 3*3*3 & [1, 1, 1] & ReLU & 117*120*6*64\\
			\hline
			Complex conv4 & 117*120*6*64 & 64 & 3*3*3 & [1, 1, 1] & ReLU & 117*120*6*64\\
			\hline
			Complex conv5 & 117*120*6*64 & 2 & 3*3*3 & [1, 1, 1] & None & 117*120*6*2\\
			\hline
			Residual & 117*120*6*2 & / & / & / & / & 117*120*6*2\\
			\hline
			Kspace data consistency & 117*120*6*2 & / & / & / & / & 117*120*6*2\\
			\hline
	\end{tabular}}
\end{table*}

The models were implemented on an Ubuntu 16.04 LTS (64-bit) operating system equipped with an Intel Xeon E5-2640 Central Processing Unit (CPU) and Tesla TITAN Xp Graphics Processing Unit (GPU, 12GB memory) in the open framework Tensorflow \cite{abadi2016tensorflow} with CUDA and CUDNN support. 

\subsubsection{Performance evaluation}
For a quantitative evaluation, mean square error (MSE), peak signal to noise ratio (PSNR) and structural similarity index (SSIM) \cite{wang2004image} were measured as follows:
\begin{equation}
\label{eq_10}
\mathrm{MSE}=
||Ref-Rec||^2_2
\end{equation}
\begin{equation}
\label{eq_11}
\mathrm{PSNR}
= 20\log_{10}\frac{\max(Ref)\sqrt{N}}{||Ref-Rec||_2}
\end{equation}
\begin{equation}
\label{eq_12}
\mathrm{SSIM}
= \boldsymbol{l}(Ref, Rec)\cdot\boldsymbol{c}(Ref, Rec)\cdot\boldsymbol{s}(Ref, Rec)
\end{equation}
where $Rec$ is the reconstructed image, $Ref$ denotes the reference image and $N$ is the total number of image pixels. The SSIM index is a multiplicative combination of the luminance term, the contrast term, and the structural term (details shown in \cite{wang2004image}).

\subsubsection{Experimental design}
To explore the effects of different components on dynamic MR reconstruction, we compare five CNN models as shown in Table \ref{models}. Specifically, we explore the effectiveness of the FDN in section III-B, where the D5C5 model and the Model 1 are compared. In section III-C and III-D, we respectively explore the impacts of the k-space loss and the spatial loss, where the Model 1 and Model 2 are compared in section III-C and the Model 1 and the Model 3 are compared in section III-D. Finally, in section III-E, the DIMENSION model, which contains all three components, will be compared with classical CS-based methods and the state-of-the-art CNN-based method to demonstrate the effectiveness of the proposed method.
\begin{table}[!t]
	\renewcommand{\arraystretch}{1.1}
	\renewcommand\tabcolsep{3pt}
	\caption{\textcolor{black}{The five models we configure to explore the effects of different components on dynamic MR reconstruction.}}
	\label{models}
	\centering
	\textcolor{black}{\begin{tabular}{|c|c|c|c|c|} \hline
			 & SDN & FDN & K-space loss & Spatial loss \\ \hline
			D5C5 & $\checkmark$ &  $\times$ & $\times$ & $\times$ \\
			\hline
			Model 1 & $\checkmark$ & $\checkmark$ & $\times$ & $\times$ \\
			\hline
			Model 2 & $\checkmark$ & $\checkmark$ & $\checkmark$ & $\times$\\
			\hline
			Model 3 & $\checkmark$ & $\checkmark$ & $\times$ & $\checkmark$\\
			\hline
			DIMENSION & $\checkmark$ & $\checkmark$ & $\checkmark$ & $\checkmark$ \\
			\hline
	\end{tabular}}
\end{table}

\subsection{Does the Frequency Domain Network Work?}
\begin{figure}[!t]
	\centering
	\subfloat{\includegraphics[width=1\linewidth]{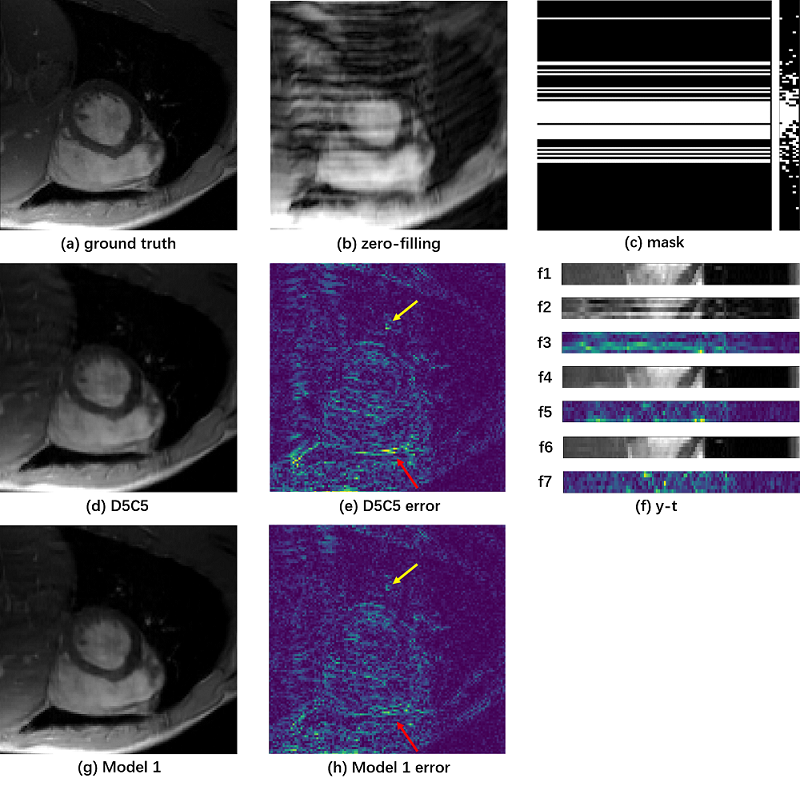}
		\label{Subresult1}}	
	\caption{The reconstruction results of the D5C5 model and the Model 1. (a) ground truth, (b) zero-filling, (c) mask and its k-t extraction,  (d) the D5C5 reconstruction, (g) the Model 1 reconstruction, (e) and (h) their corresponding error maps with display ranges [0, 0.07]. (f) Extractions of $55^{th}$ slice along y and temporal dimensions (y-t), from top (f1) to bottom (f7): (f1) ground truth, (f2) zero-filling image, (f3) error map of zero-filling, (f4) the D5C5 reconstruction, (f5) error map of the D5C5, (f6) the Model 1 reconstruction, (f7) error map of the Model 1.}
	\label{result1}
\end{figure}

To demonstrate the efficacy of the FDN, we configure the Model 1, where the weights of the k-space loss and the spatial loss are set to zeros, with the state-of-the-art CNN method D5C5 \cite{schlemper2018deep}. So the total loss function in this section is:
\begin{equation}
\label{eq_13}
\mathrm{Tloss}
= ||\textbf{S}-\hat{\textbf{S}}_4||_2^2
\end{equation}
The two models have the same amount of network parameters. And for fair comparisons, the networks’ hyperparameters are also set to be the same. The reconstruction results of the D5C5 and the Model 1 on the test datasets are shown in Fig. \ref{result1}.
 The display ranges for the error maps are $[0,\ 0.07]$. At 4-fold acceleration, one can see that the Model 1 reconstruction outperforms the D5C5 reconstruction in terms of artifacts removal and detail preservation (see the arrows in Fig. \ref{result1} (e) and (h)). To explore the reconstructed results in the direction of temporal, we show the slices along $y$ and temporal dimensions ($y-t$). From the $y-t$ images (Fig. \ref{result1} (f)), we can see that the reconstruction of the Model 1 has a smaller error map, which is consistent with the above conclusion. We also show the quantitative evaluations of the Model 1 and the D5C5 model in Table \ref{TAB}. 
 \begin{table}[!t]
 	\renewcommand{\arraystretch}{1.1}
 	\renewcommand\tabcolsep{3pt}
 	\caption{\textcolor{black}{The MSE, PSNR and SSIM of zero-filling, D5C5 and Model 1.}}
 	\label{TAB}
 	\centering
 	\textcolor{black}{\begin{tabular}{l|ccc}\hline\hline
 			Models&MSE&PSNR&SSIM\\\hline\\
 			Zero-filling&$0.004724$&$23.2146$&	$0.7637$\\\\
 			D5C5&$0.000301$&$35.1694$&$	0.9775$\\\\
 			Model 1&$\textbf{0.000176}$&$	\textbf{37.5857}$&$\textbf{0.9846}$\\\\
 			\hline\hline 
 	\end{tabular}
 \begin{tablenotes}
 	\item[1] The bolder ones mean better.
\end{tablenotes}}
 \end{table}
 One can see that the Model 1 achieves optimal quantitative evaluations over the D5C5 model (0.000128 lower in MSE, 2.4163dB higher in PSNR and 0.071 higher in SSIM). So the Model 1 is superior to the D5C5 model in both visual results and quantitative indicators. 
 
This indicates the Model 1 could effectively learn cross-domain information and improve MR reconstruction by using both frequential and spatial prior knowledge. Subsequent experiments are all constructed based on the Model 1.

\subsection{Does the K-space Loss Work?}
\begin{figure}[!t]
	\centering
	\subfloat{\includegraphics[width=1\linewidth]{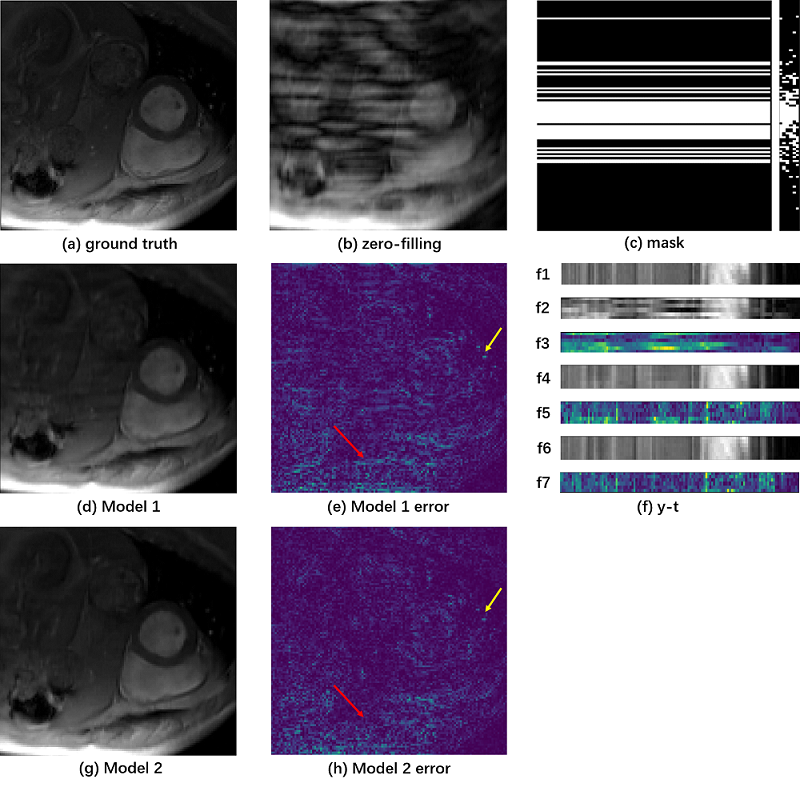}
		\label{SubKLoss}}
	
	\caption{The reconstructions of the Model 1 and the Model 2. (a) ground truth, (b) zero-filling image, (c) mask and its k-t extraction, (d) Model 1 reconstruction, (g) Model 2 reconstruction, (e) and (h) their corresponding error maps with display ranges [0, 0.07]. (f) Extractions of $55^{th}$ slice along y and temporal dimensions (y-t), from top (f1) to bottom (f7): (f1) ground truth, (f2) zero-filling image, (f3) error map of zero-filling, (f4) Model 1 reconstruction, (f5) error map of Model 1, (f6) Model 2 reconstruction, (f7) error map of Model 2.}
	\label{KLoss}
\end{figure}

This section mainly explores whether the k-space loss can improve the reconstruction results. The purpose of introducing the k-space loss is to further constrain the FDN to get a better network predicted k-space. The k-space loss is shown in Eq. \ref{eq_6}. We select the MSE between predicted k-space and fully sampled k-space as the k-space loss and the total loss function in this section is shown below:
\begin{equation}
\label{eq_14}
\mathrm{Tloss}
= ||\textbf{S}-\hat{\textbf{S}}_4||_2^2+\alpha_1||\textbf{K}_f-\textbf{K}_{\textbf{F}_{net_1}}^{DC}||_2^2
\end{equation}
where $\alpha_1$ is a hyperparameter. We refer to this model as Model 2 and choose $\alpha_1=10^{-1}$ here. The comparison results of the Model 1 and the Model 2 are shown in Fig. \ref{KLoss}.
 Obviously, the Model 2 gets better results, especially in removing artifacts. We can obtain the same conclusions from the y-t images (Fig. \ref{KLoss} (f)). The quantitative measurements can be found in Table \ref{TAB1}.
  \begin{table}[!t]
 	\renewcommand{\arraystretch}{1.1}
 	\renewcommand\tabcolsep{3pt}
 	\caption{\textcolor{black}{The MSE, PSNR and SSIM of zero-filling, Model 1 and Model 2.}}
 	\label{TAB1}
 	\centering
 	\textcolor{black}{\begin{tabular}{l|ccc}\hline\hline
 			Models&MSE&PSNR&SSIM\\\hline\\
 			Zero-filling&$0.004881$&$22.7804$&$0.8016$\\\\
 			Model 1&$0.000105$&$39.4711$&$0.9910$\\\\
 			Model 2&$\textbf{0.000061}$&$\textbf{41.8101}$&$\textbf{0.9939}$\\\\
 			\hline\hline 
 	\end{tabular}
 \begin{tablenotes}
 	\item[1] The bolder ones mean better.
\end{tablenotes}}
 \end{table}
  Introducing the k-space loss can improve the MSE, PSNR and SSIM (0.000044 lower in MSE, 2.3390dB higher in PSNR and 0.0029 higher in SSIM). Therefore, we can see that the k-space loss can effectively improve the cardiac MR reconstruction. With the joint action of KDC and k-space loss, more accurate predicted k-space can be obtained.

\subsection{Does the Spatial Loss Work?}
\begin{figure}[!t]
 	\centering
 	\subfloat{\includegraphics[width=1\linewidth]{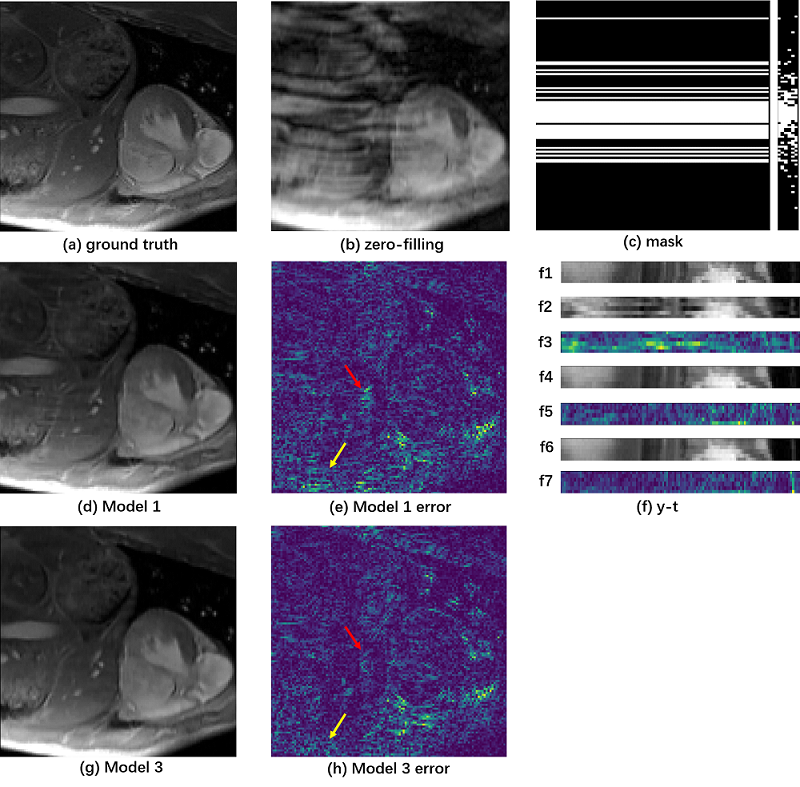}
 		\label{SubSLoss1}}
 	
 	\caption{The reconstructions of the Model 1 and the Model 3. (a) ground truth, (b) zero-filling image, (c) mask and its k-t extraction, (d) Model 1 reconstruction, (g) Model 3 reconstruction, (e) and (h) their corresponding error maps with display ranges [0, 0.07]. (f) Extractions of $55^{th}$ slice along y and temporal dimensions (y-t), from top (f1) to bottom (f7): (f1) ground truth, (f2) zero-filling image, (f3) error map of zero-filling, (f4) Model 1 reconstruction, (f5) error map of Model 1, (f6) Model 3 reconstruction, (f7) error map of Model 3.}
 	\label{SLoss1}
 \end{figure}

In this section, we demonstrate whether the spatial loss can improve the MR reconstruction. The spatial loss (in Eq. \ref{eq_7}) is to constrain the reconstruction results of different blocks with individual weights. We can regard the outputs of the shallow blocks as the preliminary reconstruction, and the output of the last block as the final reconstruction. To demonstrate the effectiveness of the spatial loss, we use the combination of the primary loss and the spatial loss as the total loss in this section:
\begin{equation}
\label{eq_15}
\mathrm{Tloss}
= ||\textbf{S}-\hat{\textbf{S}}_4||_2^2+\sum_{n=1}^{3}\beta_n||\textbf{S}-\hat{\textbf{S}}_n||_2^2
\end{equation}
Here, we refer to this model as Model 3 and choose $\beta_1=\beta_2=\beta_3=10^3$. Fig. \ref{SLoss1} shows the reconstruction results of the Model 1 and the Model 3. 
 We can clearly see that the Model 3 has fewer artifacts and retains details better from the error maps (as shown by the red and yellow arrows in Fig. \ref{SLoss1} (e, h, f5, f7).
   \begin{table}[!t]
 	\renewcommand{\arraystretch}{1.1}
 	\renewcommand\tabcolsep{3pt}
 	\caption{\textcolor{black}{The MSE, PSNR and SSIM of zero-filling, Model 1 and Model 3.}}
 	\label{TAB2}
 	\centering
 	\textcolor{black}{\begin{tabular}{l|ccc}\hline\hline
 			Models&MSE&PSNR&SSIM\\\hline\\
 			Zero-filling&$0.004211$&$23.7471$&$0.8208$\\\\
 			Model 1&$0.000261$&$35.8304$&$0.9738$\\\\
 			Model 3&$\textbf{0.000214}$&$\textbf{36.6892}$&$\textbf{0.9764}$\\\\
 			\hline\hline 
 		\end{tabular}
 		\begin{tablenotes}
 			\item[1] The bolder ones mean better.
 	\end{tablenotes}}
 \end{table}
 The quantitative measurements can be found in Table \ref{TAB2}, from which we can see that the spatial loss can effectively improve the MSE, PSNR and SSIM (0.000047 lower in MSE, 0.8588dB higher in PSNR and 0.0026 higher in SSIM). Therefore, we can demonstrate that the spatial loss can further improve the cardiac MR reconstruction.

\subsection{Comparison to the State-of-the-art Methods}
Based on the above sections, the network we used is the DIMENSION model (as shown in Fig. \ref{networks}) and the multi-supervised loss we selected is 
\begin{equation}
\label{eq_16}
\mathrm{Tloss}
= ||\textbf{S}-\hat{\textbf{S}}_4||_2^2+\sum_{n=1}^{3}10^3||\textbf{S}-\hat{\textbf{S}}_n||_2^2+10^{-1}||\textbf{K}_f-\textbf{K}_{\textbf{F}_{net_1}}^{DC}||_2^2
\end{equation}

 In this section, we will demonstrate the effectiveness of the proposed method by comparing it with the traditional CS (k-t FOCUSS \cite{jung2007improved}, k-t SLR \cite{lingala2011accelerated}, S+L \cite{otazo2015low}) methods and the state-of-the-art CNN method (D5C5 \cite{schlemper2018deep}). We did not compare our method with CRNN \cite{qin2018convolutional} since the CRNN model and the DC-CNN model share the comment property such as their networks are built in the spatial domain with a single supervised loss function. Without loss of generality, the validity of the proposed network structures and the multi-supervised learning can be demonstrated by only comparing with the DC-CNN model. And the CRNN code is not available. 
 
\begin{figure*}[!t]
	\centering
	\subfloat{\includegraphics[width=0.8\linewidth]{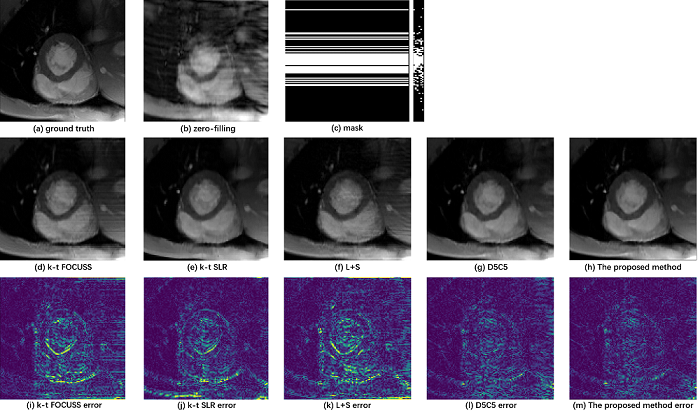}
		\label{Subresults}}	
	\caption{The comparison of cardiac MR reconstructions from different methods (k-t FOCUSS, k-t SLR, L+S, D5C5, the proposed method). (a) ground truth, (b) zero-filling image, (c) mask and its k-t extraction (d) k-t FOCUSS reconstruction, (e) k-t SLR reconstruction, (f) L+S reconstruction, (g) D5C5 reconstruction, (h) the proposed method reconstruction; (i), (j), (k), (l) and (m) their corresponding error maps with display ranges [0, 0.07].}
	\label{results}
\end{figure*}

 All the CS-based methods select their single-channel versions. For fair comparisons, we adjust the parameters of the CS-MRI methods to their best performances. The reconstruction results at 4-fold acceleration from these methods are shown in Fig. \ref{results}.
 The k-t SLR removes artifacts better than the k-t FOCUSS and S+L. However, these three CS-based methods lose more structural details than the CNN-based methods. Compared with the D5C5 method, the proposed method can not only retain more details, but also remove more artifacts. 

We show the evaluation indexes in Table \ref{TAB5}. 
\begin{table}[!t]
	\renewcommand{\arraystretch}{1.1}
	\renewcommand\tabcolsep{3pt}
	\caption{\textcolor{black}{The MSE, PSNR, SSIM and running time of zero-filling, k-t FOCUSS, k-t SLR, D5C5 and the proposed method.}}
	\label{TAB5}
	\centering
	\textcolor{black}{\begin{tabular}{l|ccc}\hline\hline
			Models&MSE&PSNR&SSIM\\\hline\\
			Zero-filling&$0.006003$&$22.2162$&$	0.5346$\\\\
			k-t FOCUSS&$0.000308$&$35.1201$&$0.9162$\\\\
			k-t SLR&$0.000157$&$38.0534$&$0.9532$\\\\
			S+L&$0.000333$&$34.7778$&$0.9616$\\\\
			D5C5&$0.000121$&$39.1788$&$0.9548$\\\\
			The proposed&$\textbf{0.000086}$&$\textbf{40.6256}$&$\textbf{0.9655}$\\\\
			\hline\hline 
		\end{tabular}
		\begin{tablenotes}
			\item[1] The bolder ones mean better.
	\end{tablenotes}}
\end{table}
Note that the CNN-based methods outperform the CS-based methods in all three performance indexes. We observe the proposed DIMENSION model achieves the optimal performance in MSE, PSNR and SSIM indexes among all the methods. The reconstruction time of different methods is shown in Table \ref{Time}. And for fair comparsisons, all methods are implemented on the same CPU Intel Xeon E5-2640. It can be seen that the reconstruction time of the CNN-based methods is significantly shorter than the CS-based methods.
\begin{table*}[!t]
	\renewcommand{\arraystretch}{1.1}
	\renewcommand\tabcolsep{3pt}
	\caption{\textcolor{black}{The reconstruction time of k-t SLR, L+S, D5C5, the proposed methods.}}
	\label{Time}
	\centering
	\textcolor{black}{\begin{tabular}{|l|c|c|c|c|}\hline
			Methods&k-t SLR&L+S&D5C5&The proposed\\\hline
			Running time/s&$14.7249$&$17.3808$&$4.9318$&$\textbf{4.8845}$\\
			\hline
		\end{tabular}
		\begin{tablenotes}
			\item[1] The bolder ones mean better.
	\end{tablenotes}}
\end{table*}

\begin{figure*}[!t]
	\centering
	\subfloat{\includegraphics[width=0.7\linewidth]{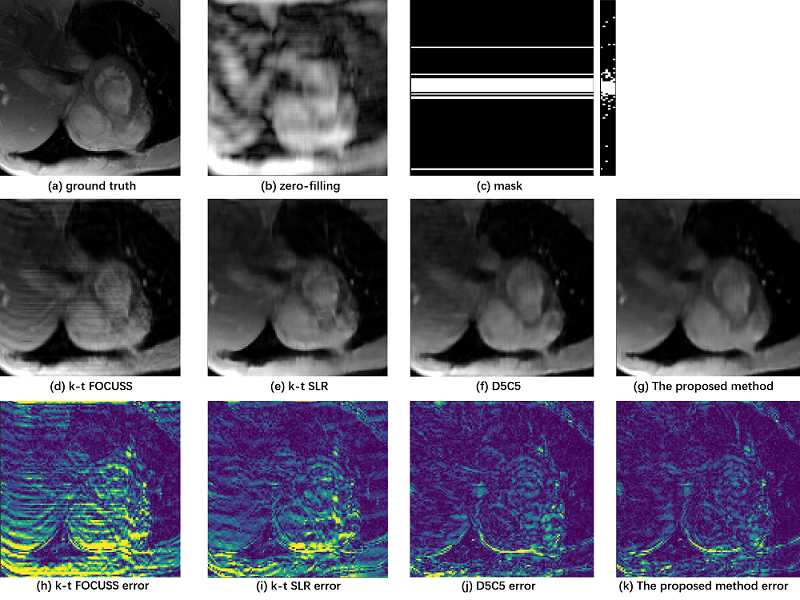}
		\label{Subresults_discussion}}	
	\caption{The comparison of cardiac MR reconstructions from different methods (k-t FOCUSS, k-t SLR, D5C5, the proposed method) at 8-fold acceleration. (a) ground truth, (b) zero-filling image, (c) mask and its k-t extraction, (d) k-t FOCUSS reconstruction, (e) k-t SLR reconstruction, (f) D5C5 reconstruction, (g) the proposed method reconstruction; (h), (i), (j), (k) their corresponding error maps with display ranges [0, 0.07].}
	\label{results_discussion}
\end{figure*}

At 8-fold acceleration, the reconstruction results of different methods are shown in Fig. \ref{results_discussion}. At high acceleration factors, the single-channel CS-based methods have difficulties in reconstructing high quality images. However, the CNN-based methods can still get better reconstruction results. Compared with the D5C5 method, the proposed method can not only retain the details, but also remove the artifacts better. The improvements of CNN-based methods in reconstruction are more obvious at high acceleration factors. The evaluation indexes are shown in Table. \ref{TAB6}. We observe that the proposed method achieves the optimal performances in MSE, PSNR and SSIM indexes among all the methods. 
\begin{table}[!t]
	\renewcommand{\arraystretch}{1.1}
	\renewcommand\tabcolsep{3pt}
	\caption{\textcolor{black}{The MSE, PSNR, SSIM and running time of zero-filling, k-t FOCUSS, k-t SLR, D5C5 and the proposed method.}}
	\label{TAB6}
	\centering
	\textcolor{black}{\begin{tabular}{l|ccc}\hline\hline
			Models&MSE&PSNR&SSIM\\\hline\\
			Zero-filling&$0.00380$&$20.4509$&$0.6599$\\\\
			k-t FOCUSS&$0.001159$&$21.5761$&$0.8539$\\\\
			k-t SLR&$0.000449$&$25.6889$&$0.9103$\\\\
			D5C5&$0.000489$&$31.6137$&$0.9649$\\\\
			DIMENSION&$\textbf{0.000381}$&$\textbf{32.7019}$&$\textbf{0.9690}$\\\\
			\hline\hline 
		\end{tabular}
		\begin{tablenotes}
			\item[1] The bolder ones mean better.
	\end{tablenotes}}
\end{table}

\section{DISCUSSION} 

\subsection{Hyper-parameters Selection}
The weights selection of the k-space loss and the spatial loss influence the reconstruction \cite{lecun2015deep}. In this section, we will discussion the selections of $\alpha_1$ and $\beta_n,\ n=1, 2, 3$.
\subsubsection{The Selection of $\alpha_1$}
\begin{figure}[!t]
	\centering
	\subfloat{\includegraphics[width=0.9\linewidth]{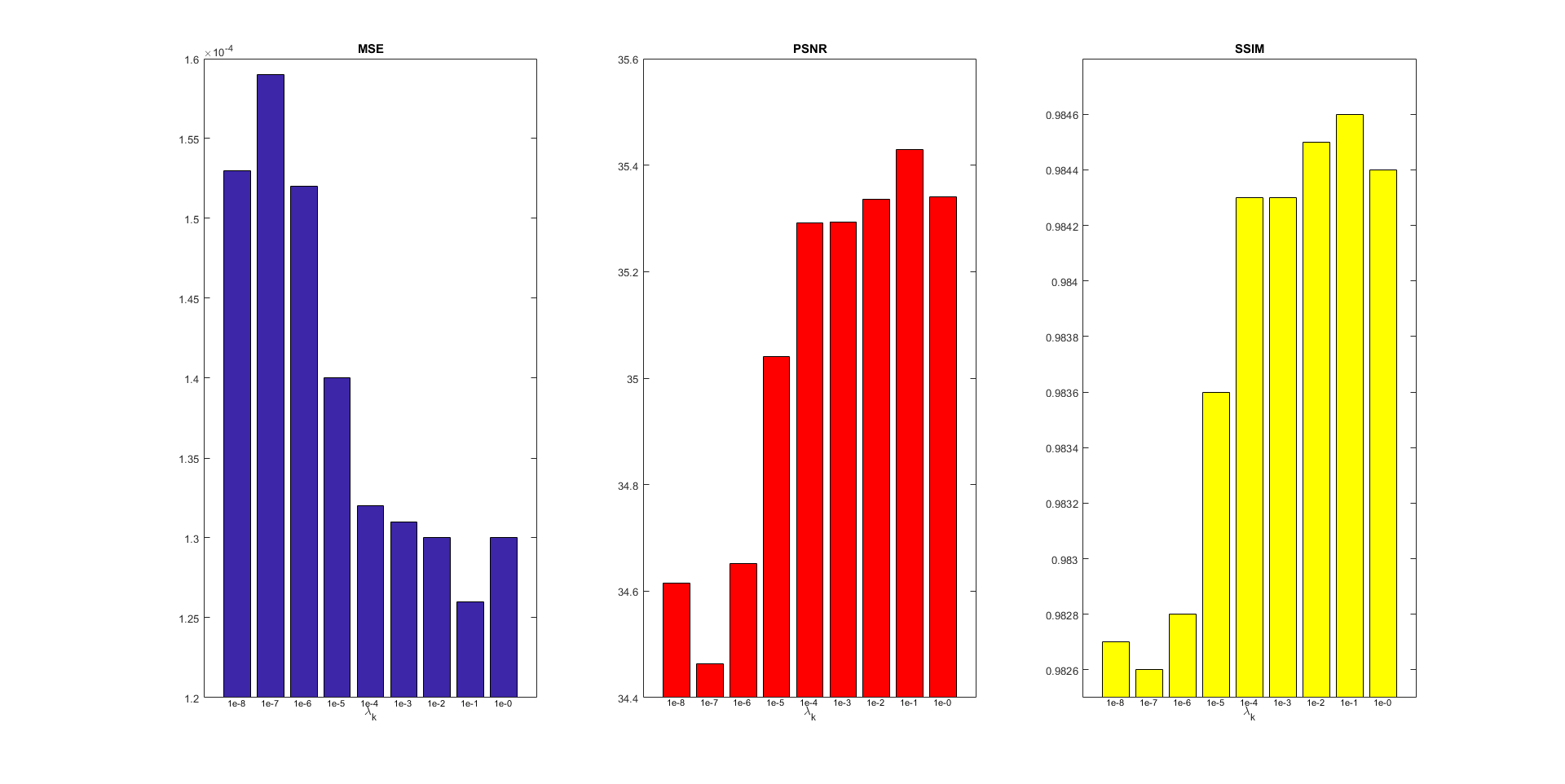}
		\label{Sublambda_k}}	
	\caption{The histograms of the average MSE, PSNR, SSIM on the test set with different $\alpha_1$. Here, $\alpha_1$ goes from $10^{-8}$ to $10^{-1}$}
	\label{lambda_k}
\end{figure}

In order to find an appropriate $\alpha_1$, we trained a series of Model 2 with different $\alpha_1$ and then tested them on the same test set. The average MSE, PSNR, SSIM on the test set are shown in Fig. \ref{lambda_k}.
We can observe that when $\alpha_1=10^{-1}$, the Model 2 gets the lowest MSE, the highest PSNR and the highest SSIM. So we pick $10^{-1}$ as the appropriate value for $\alpha_1$.

\subsubsection{The Selection of $\beta_n$}
 There are two ways to select $\beta_n$:  the one is to take the same values for these three weights. And the other is to take increasing or decreasing values for these three weights.

\begin{figure}[!t]
	\centering
	\subfloat{\includegraphics[width=1\linewidth]{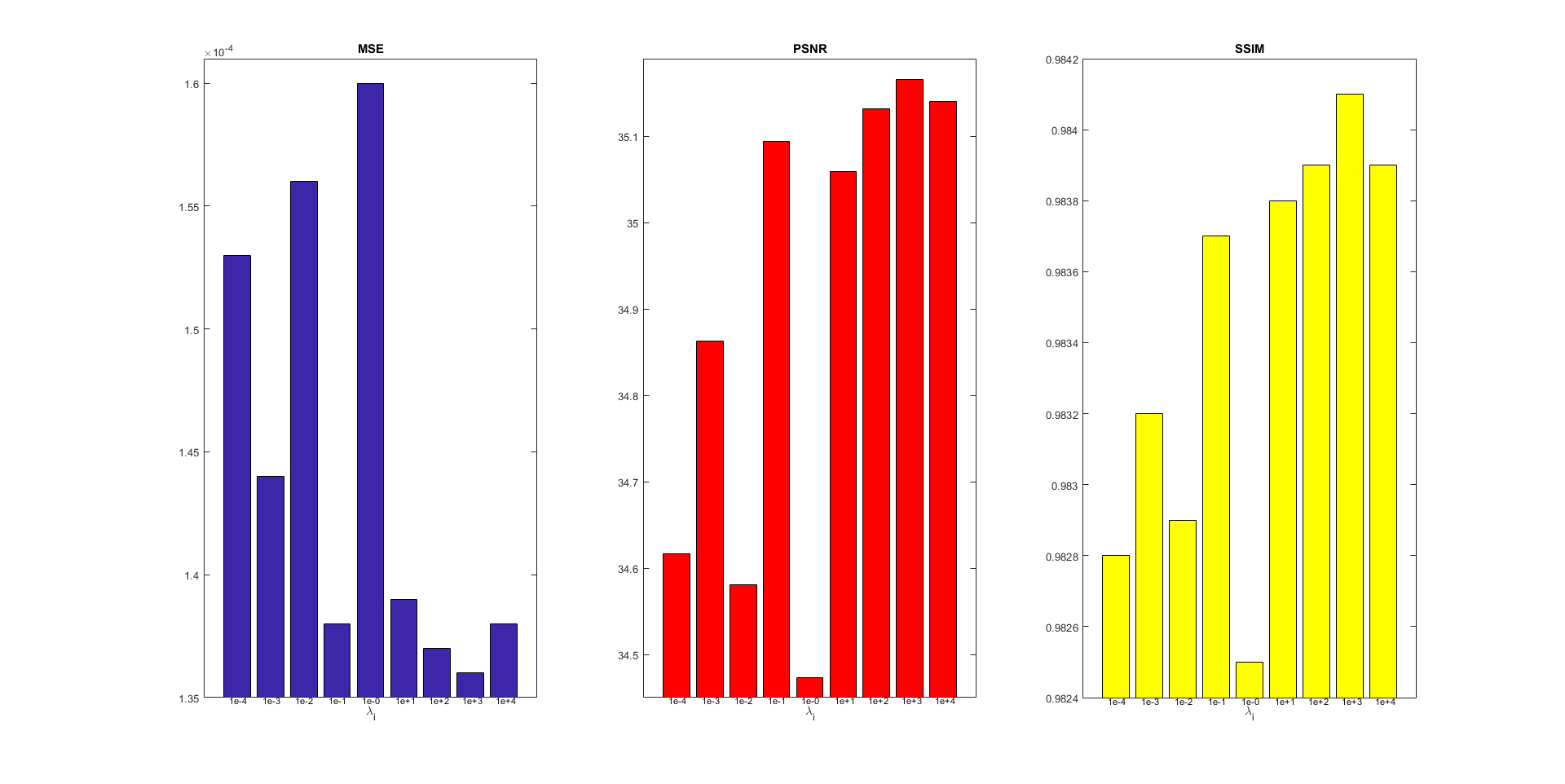}
		\label{Sublambda_i}}
	
	\caption{The histograms of the average MSE, PSNR, SSIM on the test set with different $\beta_n, n=1,2,3$. Here, $\beta_1=\beta_2=\beta_3$ and equal to $10^{-4}, 10^{-3}, \dots, 10^{3}, 10^{4}$}
	\label{lambda_i}
\end{figure}

Firstly, we consider the case when each $\beta_n$ takes the same value. We trained a series of Model 3, where $\beta_n,\ n=1,2,3$, in turn, equal to $10^{-4}, 10^{-3}, \dots, 10^3, 10^4$. Then all these models were tested on the same test set. The quantitative results are shown in Fig. \ref{lambda_i}.
We can see that $\beta_1=\beta_2=\beta_3=10^3$ has the best performance on the test set with the lowest MSE, the highest PSNR and the highest SSIM. We refer to the Model 3 with $\beta_1=\beta_2=\beta_3=10^3$ as DIMENSION-SLoss1.

\begin{figure}[!t]
	\centering
	\subfloat{\includegraphics[width=1\linewidth]{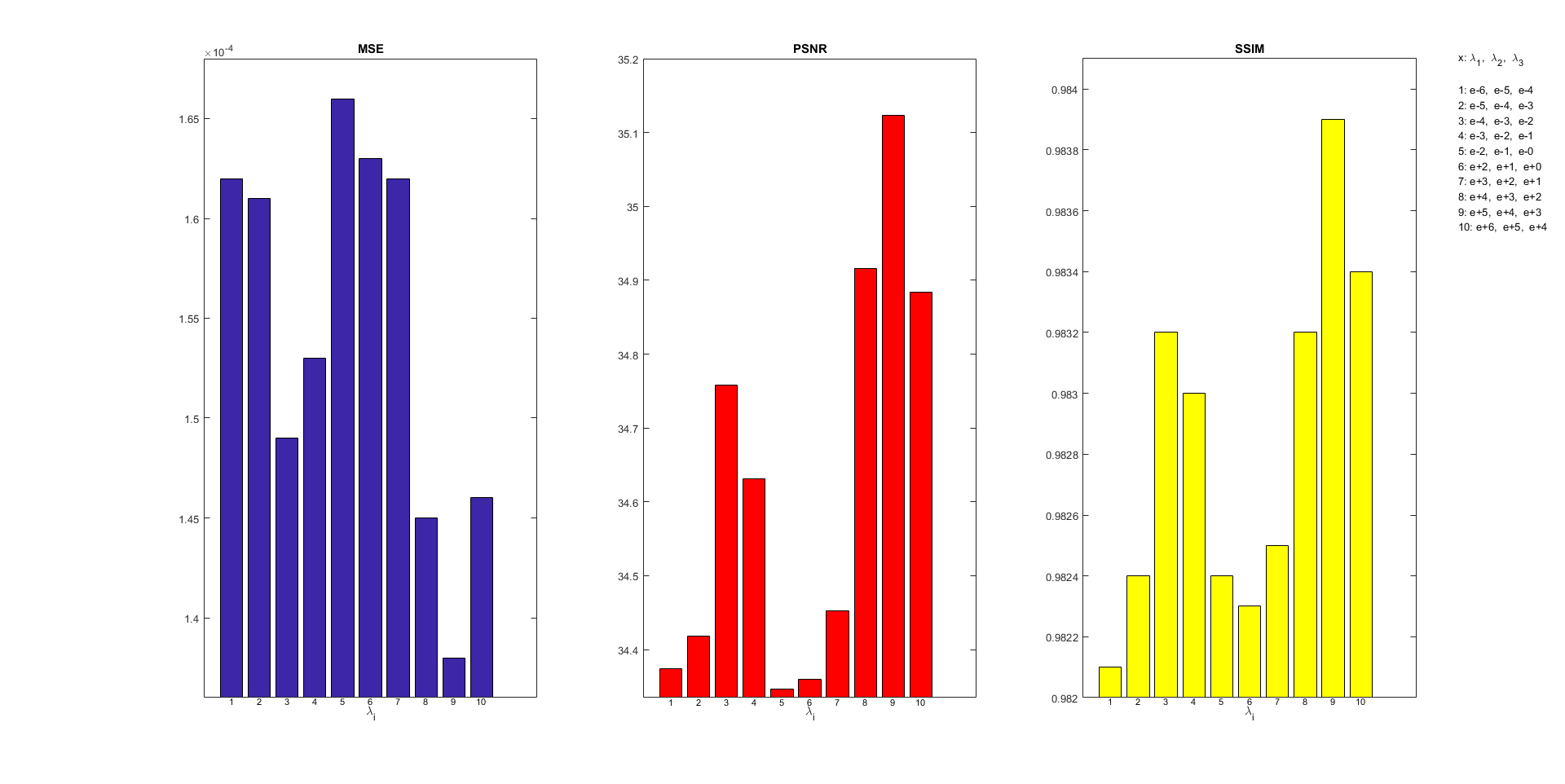}
		\label{Sublambda_i2}}
	
	\caption{The histograms of the average MSE, PSNR, SSIM on the test set with different $\beta_n, n=1,2,3$, which shown in Table \ref{Case2}.}
	\label{lambda_i2}
\end{figure}

Secondly, we consider the case when these three weights take increasing or decreasing values. The values of $\beta_n,\ n=1,2,3$ are shown in Table \ref{Case2}. 
\begin{table*}[!t]
	\renewcommand{\arraystretch}{1.1}
	\renewcommand\tabcolsep{3pt}
	\caption{\textcolor{black}{Ten cases of $\beta_n, n=1,2,3$} taking different values.}
	\label{Case2}
	\centering
	\textcolor{black}{\begin{tabular}{|c|c|c|c|c|c|c|c|c|c|c|} \hline
			$\beta_n$ & Case 1 & Case 2 & Case 3 & Case 4 & Case 5  & Case 6 & Case 7 & Case 8 & Case 9 & Case 10\\ \hline
			$\beta_1$ & $10^{-6}$ & $10^{-5}$ & $10^{-4}$ & $10^{-3}$ & $10^{-2}$ & $10^{2}$ & $10^{3}$ & $10^{4}$ & $10^{5}$ & $10^{6}$\\
			\hline
			$\beta_2$ & $10^{-5}$ & $10^{-4}$ & $10^{-3}$ & $10^{-2}$ & $10^{-1}$ & $10^{1}$ & $10^{2}$ & $10^{3}$ &  $10^{4}$ & $10^{5}$\\
			\hline
			$\beta_3$ & $10^{-4}$ & $10^{-3}$ & $10^{-2}$ & $10^{-1}$ & $10^{-0}$ & $10^{0}$ & $10^{1}$ & $10^{2}$ & $10^{3}$ & $10^{4}$\\
			\hline
	\end{tabular}}
\end{table*}
We have considered a total of 10 cases, where $\beta_n,\ n=1,2,3$ are increasing in the first five and decreasing in the last five. The quantitative results are shown in Fig. \ref{lambda_i2}.
Obviously, we find that the ninth case gets the best quantization results, where MSE is the lowest, PSNR and SSIM are the highest. So, we choose $\beta_1=10^5,\ \beta_2=10^4,\ \beta_3=10^3$ as our preferred weights. We refer to the Model 3 with $\beta_1=10^5,\ \beta_2=10^4,\ \beta_3=10^3$ as DIMENSION-SLoss2.

\begin{figure}[!t]
	\centering
	\subfloat{\includegraphics[width=1\linewidth]{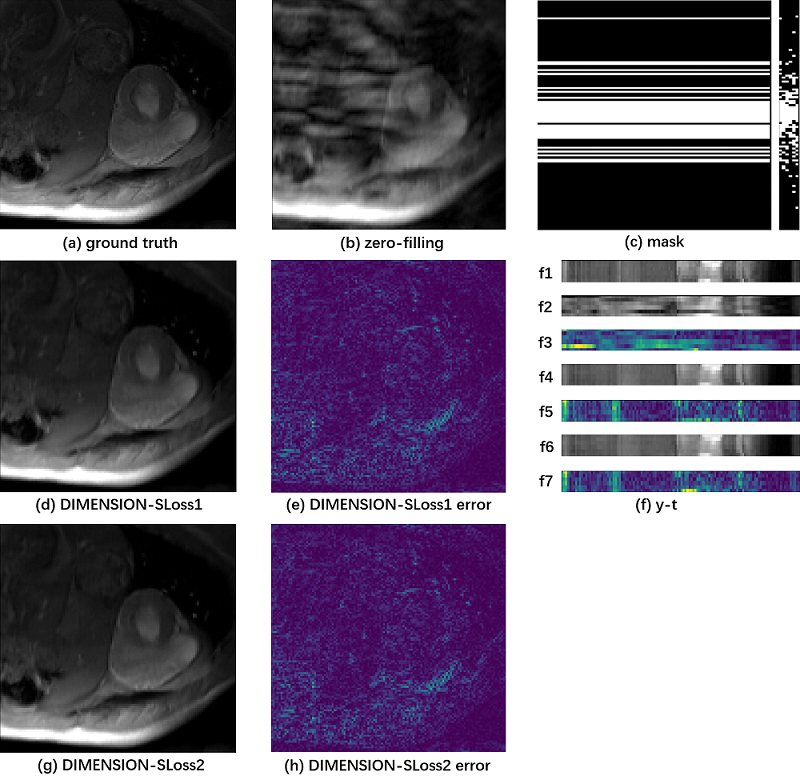}
		\label{SubSLoss1and2}}
	
	\caption{The reconstructions of the DIMENSION-SLoss1 model and the DIMENSION-SLoss2 model. (a) ground truth, (b) zero-filling image, (c) mask and its k-t extraction, (d) DIMENSION-SLoss1 reconstruction, (g) DIMENSION-SLoss2 reconstruction, (e) and (h) their corresponding error maps with display ranges [0, 0.07]. (f) Extractions of $55^{th}$ slice along y and temporal dimensions (y-t), from top (f1) to bottom (f7): (f1) ground truth, (f2) zero-filling image, (f3) error map of zero-filling, (f4) DIMENSION-SLoss1 reconstruction, (f5) error map of DIMENSION-SLoss1, (f6) DIMENSION-SLoss2 reconstruction, (f7) error map of DIMENSION-SLoss2.}
	\label{SLoss1and2}
\end{figure}

Then, it is natural to think about which of these two ways to select $\beta_n$ works better. The quantitative and reconstruction results of the DIMENSION-SLoss1 model and the DIMENSION-SLoss2 model are respectively shown in Fig. \ref{SLoss1and2}
and Table \ref{TAB4}.
\begin{table}[!t]
	\renewcommand{\arraystretch}{1.1}
	\renewcommand\tabcolsep{3pt}
	\caption{\textcolor{black}{The MSE, PSNR and SSIM of zero-filling, DIMENSION-SLoss1 and DIMENSION-SLoss2.}}
	\label{TAB4}
	\centering
	\textcolor{black}{\begin{tabular}{l|ccc}\hline\hline
			Models&MSE&PSNR&SSIM\\\hline\\
			Zero-filling&$0.004297$&$23.6684$&$0.8169$\\\\
			DIMENSION-SLoss1&$\textbf{0.000083}$&$\textbf{40.8160}$&$\textbf{0.9918}$\\\\
			DIMENSION-SLoss2&$0.000094$&$40.2876$&$	0.9913$\\\\
			\hline\hline 
		\end{tabular}
		\begin{tablenotes}
			\item[1] The bolder ones mean better.
	\end{tablenotes}}
\end{table}
The two models have similar reconstructed results, but the quantization results of DIMENSION-SLoss1 are slightly improved. So we choose DIMENSION-SLoss1 as the final Model 3. 

\subsection{The Limitations of the Proposed Work}
Although our proposed method achieves the improved reconstruction results for dynamic MR imaging compared to other methods, there is still a certain degree of smooth in the reconstructed images at high acceleration factors. Part of the reasons may be the loss functions used in this work. The MSE loss functions only indicate the mean square information between the reconstructed image and the ground truth and cannot perceive the image structure information. DAGAN \cite{yang2018dagan} couples an adversarial loss with an innovative content loss to reconstruct CS-MRI, which could preserve perceptual image details. This inspired us to use different loss functions related to structural information in the future works. Furthermore, network structures could have other options to improve the reconstruction, which are going to be explored and investigated. For example, Dense networks \cite{zhang2018residual} may be utilized to make full use of hierarchical features. 

\section{Conclusion and Outlook}
 In this work, we propose a dynamic MR imaging method with both k-space and spatial prior knowledge integrated via multi-supervised network training, dubbed as DIMENSION. Our contributions are mainly reflected in the cross-domain network structures and the multi-supervised loss function strategy. The cross-domain network structures help us to make better use of both frequential domain and spatial domain prior knowledge. The multi-supervised loss function strategy avoids the single-supervised training and provides different levels of supervision for the entire network. Specifically, we introduce the k-space loss, which can force the predicted k-space from the frequency domain network to be as close as possible to the fully sampled k-space. We also introduce the spatial loss into the spatial domain network, which can constrain the reconstruction results at different levels. We compared the proposed approach with k-t FOCUSS, k-t SLR, L+S and the state-of-the-art CNN-based method. Experimental results show that the proposed network structure and loss function strategy can improve dynamic MR reconstruction accuracy with shorter time. In future works, perceived loss functions and more advanced network structures may be studied and used.

\ifCLASSOPTIONcaptionsoff
  \newpage
\fi



\bibliographystyle{IEEEtran}
\bibliography{IEEEabrv,DIMENSION}
\end{document}